\documentclass{article} 

\usepackage{microtype}
\usepackage{epsfig}
\usepackage{graphicx}
\usepackage{wrapfig}
\usepackage{caption}
\usepackage{subcaption}

\usepackage{multirow}
\usepackage{rotating}
\usepackage{booktabs}

\usepackage{hyperref}

\newcommand{\mcrot}[4]{\multicolumn{#1}{#2}{\rlap{\rotatebox{#3}{#4}~}}}

\usepackage{eso-pic}
\usepackage{xspace}
\makeatletter
\DeclareRobustCommand\onedot{\futurelet\@let@token\@onedot}
\def\@onedot{\ifx\@let@token.\else.\null\fi\xspace}
\def\eg{\emph{e.g}\onedot} 
\def\ie{\emph{i.e}\onedot}

\def\etc{\emph{etc}\onedot} \def\vs{\emph{vs}\onedot}

\makeatother

\usepackage[accepted]{icml2020}

\icmltitlerunning{Probing Emergent Semantics in Predictive Agents via Question Answering}

\begin{document}

\twocolumn[
\icmltitle{Probing Emergent Semantics in Predictive Agents via Question Answering}



\icmlsetsymbol{equal}{*}

\begin{icmlauthorlist}
\icmlauthor{Abhishek Das*}{to}
\icmlauthor{Federico Carnevale*}{at}
\icmlauthor{Hamza Merzic}{at}
\icmlauthor{Laura Rimell}{at}
\icmlauthor{Rosalia Schneider}{at}
\icmlauthor{Josh Abramson}{at}
\icmlauthor{Alden Hung}{at}
\icmlauthor{Arun Ahuja}{at}
\icmlauthor{Stephen Clark}{at}
\icmlauthor{Gregory Wayne}{at}
\icmlauthor{Felix Hill}{at}
\end{icmlauthorlist}

\icmlaffiliation{to}{Georgia Institute of Technology}
\icmlaffiliation{at}{DeepMind}

\icmlcorrespondingauthor{}{felixhill@google.com}

\icmlkeywords{Machine Learning, ICML}

\vskip 0.3in
]



\printAffiliationsAndNotice{\icmlEqualContribution} 

\begin{abstract}
    Recent work has shown how predictive modeling can endow
    agents with rich knowledge of their surroundings, improving
    their ability to act in complex environments. We propose
    question-answering as a general paradigm to decode and
    understand the representations that such agents develop, applying
    our method to two recent approaches to predictive modeling --
    action-conditional CPC~\citep{guo_arxiv18} and SimCore~\citep{gregor_neurips19}.
    After training agents with these predictive objectives in a
    visually-rich, $3$D environment with an assortment of objects, colors, shapes,
    and spatial configurations, we probe their internal state representations
    with synthetic (English) questions, without backpropagating gradients
    from the question-answering decoder into the agent. The performance of
    different agents when probed this way reveals that they learn to
    encode factual, and seemingly compositional, information about
    objects, properties and spatial relations from their physical environment.
    Our approach is intuitive, \ie~humans can easily interpret responses of the model
    as opposed to inspecting continuous vectors, and
    model-agnostic, \ie applicable to any modeling approach.
    By revealing the implicit knowledge of objects, quantities, properties and relations
    acquired by agents as they learn, \emph{question-conditional agent probing} can
    stimulate the design and development of stronger
    predictive learning objectives.
\end{abstract}

\vspace{-20pt}
\section{Introduction}

Since the time of Plato, philosophers have considered the apparent distinction between ``knowing how" (procedural knowledge or skills) and ``knowing what" (propositional knowledge or facts). It is uncontroversial that deep reinforcement learning (RL) agents can effectively acquire procedural knowledge as they learn to play games or solve tasks. Such knowledge might manifest in an ability to find all of the green apples in a room, or to climb all of the ladders while avoiding snakes. However, the capacity of such agents to acquire factual knowledge about their surroundings -- of the sort that can be readily hard-coded in symbolic form in classical AI -- is far from established. Thus, even if an agent successfully climbs ladders and avoids snakes, we have no certainty that it `knows' that ladders are brown, that there are five snakes nearby, or that the agent is currently in the middle of a three-level tower with one ladder left to climb.

The acquisition of knowledge about objects, properties, relations and quantities by learning-based agents is desirable for several reasons. First, such knowledge should ultimately complement procedural knowledge when forming plans that enable execution of complex, multi-stage cognitive tasks. Second, there seems (to philosophers at least) to be something fundamentally human about having knowledge of facts or propositions~\cite{stich1979animals}. If one of the goals of AI is to build machines that can engage with, and exhibit convincing intelligence to, human users (\eg justifying their behaviour so humans understand/trust them), then a need for uncovering and measuring such knowledge in learning-based agents will inevitably arise.

\begin{figure*}[t]
    \centering
    \includegraphics[width=0.75\textwidth]{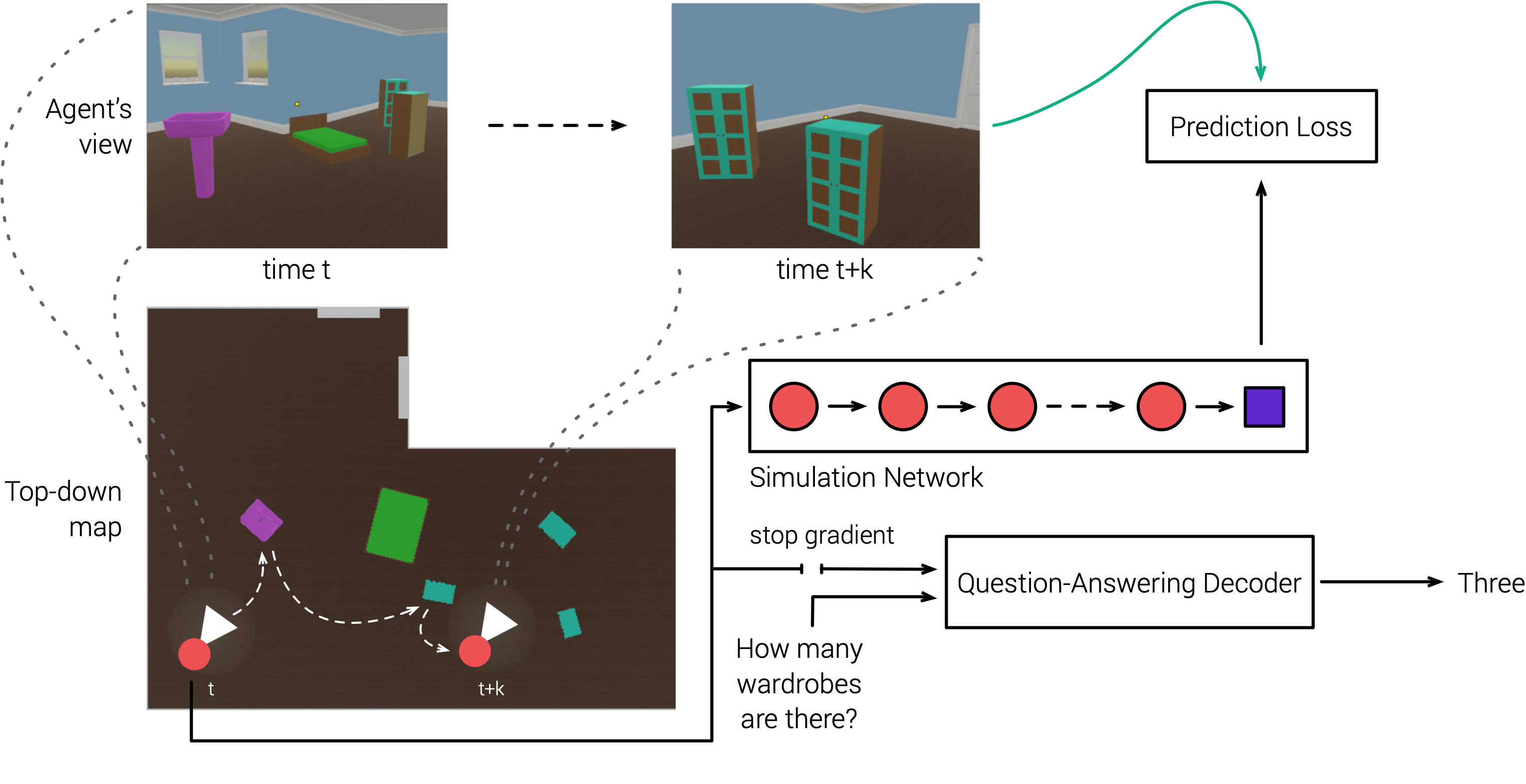}
    \vspace{-15pt}
    \caption{We train predictive agents to explore a visually-rich 3D environment
    with an assortment of objects of different shapes, colors and sizes.
    As the agent navigates (trajectory shown in white on the top-down map),
    an auxiliary network learns to simulate representations
    of future observations (labeled `Simulation Network') $k$ steps into the future,
    self-supervised by a loss against the ground-truth egocentric observation at $t+k$. Simultaneously, another decoder network is trained to extract answers to a variety of questions about the environment, conditioned on the agent's internal state but without affecting it (notice `stop gradient' -- gradients from the QA decoder are not backpropagated into the agent). We use this question-answering paradigm to decode and understand the internal representations that such agents develop. Note that the top-down map is only shown for
    illustration and not available to the agent.}
    \label{fig:teaser}
    \vspace{-15pt}
\end{figure*}

Here, we propose the question-conditional probing of agent internal states as a
means to study and quantify the knowledge about objects, properties, relations and quantities encoded in the internal representations of neural-network-based agents. Couching an analysis of such knowledge in terms of question-answering has several pragmatic advantages. First, question-answering provides a general purpose method for agent-analysis and an intuitive investigative tool for humans -- one can simply \emph{ask} an agent what it knows about its environment and get an answer back, without having to
inspect internal activations. Second, the space of questions is
essentially open-ended --  we can pose arbitrarily complex questions to an
agent, enabling a comprehensive analysis of the current state of its propositional knowledge.
Question-answering has previously been studied in
textual~\citep{rajpurkar_emnlp16,rajpurkar_acl18}, visual~\citep{malinowski_nips14,antol_iccv15,visdial}
and embodied~\citep{gordon_cvpr18,embodiedqa} settings.
Crucially, however, these systems are trained end-to-end for the goal of answering questions.
Here, we utilize question-answering simply to probe an agent's internal
representation, without backpropagating gradients from the question-answering
decoder into the agent. That is, we view question-answering as a general purpose
(conditional) decoder of environmental information designed to assist
the development of agents by revealing the extent (and limits) of their knowledge.

Many techniques have been proposed for endowing agents with
general (\ie task-agnostic) knowledge, based on both hard-coding and learning. Here, we specifically focus on the effect of self-supervised predictive modeling -- a learning-based approach -- on the acquisition of propositional knowledge. Inspired by
learning in humans~\citep{elman_1990,rao_nature99,clark16,hohwy13}, predictive modeling,~\ie predicting future sensory observations, has emerged
as a powerful method to learn general-purpose neural network
representations~\citep{elias_1955,atal_bell70,schmidhuber_1991,schaul_ijcai13,schaul_icml15,silver_icml17,wayne_arxiv2018, guo_arxiv18,gregor_neurips19,recanatesi_biorxiv19}.
These representations can be learned while exploring in and interacting with
an environment in a task-agnostic manner, and later exploited for goal-directed behavior.

We evaluate predictive \vs non-predictive agents (both trained for exploration) on our question-answering testbed
to investigate how much knowldge of object shapes, quantities, and spatial relations they acquire \emph{solely by egocentric prediction}.
The set includes a mix of questions that can plausibly be answered from a single observation or a few consecutive observations, and those that require the agent to integrate  global knowledge of its entire surroundings.

Concretely, we make the following contributions:
\vspace{-10pt}

\begin{itemize}
\itemsep0.2em

\item In a visually-rich $3$D room environment developed in the Unity
    engine, we develop a set of questions designed to probe a diverse body of factual knowledge about the environment -- from identifying shapes and colors
    (`What shape is the red object?') to counting (`How many blue objects
    are there?') to spatial relations (`What is the color of the chair
    near the table?'), exhaustive search (`Is there a cushion?'), and
    comparisons (`Are there the same number of tables as chairs?').

\item We train RL agents augmented with
    predictive loss functions -- 1) action-conditional CPC~\citep{guo_arxiv18}
    and 2) SimCore~\citep{gregor_neurips19} --
    for an exploration task and analyze the internal representations they develop by
    decoding answers to our suite of questions.
    Crucially, the QA decoder is trained independent of the predictive agent and
    we find that QA performance is indicative of the agent's ability to capture
    global environment structure and semantics \emph{solely through egocentric prediction}.
    We compare these predictive agents to strong non-predictive LSTM baselines
    as well as to an agent that is explicitly optimized for the
    question-answering task.
    \vspace{-5pt}
\item We establish generality of the encoded knowledge
    by testing zero-shot generalization of a trained QA decoder to compositionally
    novel questions (unseen combinations of seen attributes), suggesting a degree of compositionality in the internal representations captured by predictive agents.

\end{itemize}
\vspace{-10pt}
\section{Background and related work}
\label{sec:related}

Our work builds on studies of predictive modeling and auxiliary objectives in reinforcement learning as well as grounded language learning and embodied question answering.

\textbf{Propositional knowledge} is knowledge that a statement, expressed in natural or formal language, is true~\citep{truncellito2007epistemology}. Since at least Plato, epistemologist philosophers have contrasted propositional knowledge with \emph{procedural knowledge} (knowledge of how to do something), and some (but not all) distinguish this from \emph{perceptual knowledge} (knowledge obtained by the senses that cannot be translated into a proposition)~\citep{dretske1995meaningful}. An ability to exhibit this sort of knowledge in a convincing way is likely to be crucial for the long-term goal of having agents achieve satisfying interactions with humans, since an agent that cannot express its knowledge and beliefs in human-interpretable form may struggle to earn the trust of users.

\textbf{Predictive modeling and auxiliary loss functions in RL}.
The power of predictive modeling for representation learning has been known
since at least the seminal work of~\cite{elman_1990} on emergent language structures.
More recent examples include Word2Vec~\citep{mikolov_arxiv13}, Skip-Thought vectors~\citep{kiros_nips15},
and BERT~\citep{devlin_naacl19} in language, while in vision similar principles
have been applied to context prediction~\citep{doersch_iccv15,noroozi_eccv16},
unsupervised tracking~\citep{wang_iccv15}, inpainting~\citep{pathak_cvpr16} and
colorization~\citep{zhang_eccv16}. More related to us is
the use of such techniques in designing auxiliary loss functions for training
model-free RL agents, such as successor representations~\citep{dayan_nc93,zhu_iccv17},
value and reward prediction~\citep{jaderberg_arxiv16, hermann_arxiv17,wayne_arxiv2018},
contrastive predictive coding (CPC)~\citep{oord_arxiv18,guo_arxiv18},
and SimCore~\citep{gregor_neurips19}.

\begin{table*}[t!]
    \caption{QA task templates. In every episode,
      objects and their configurations are randomly generated, and these
      templates get translated to QA pairs for all unambiguous
      {\small $<${\tt shape, color}$>$} combinations. There are
      $50$ {\tt shape}s and $10$ {\tt color}s in total.
      See~\ref{sec:supp_environment} for details.
      }
    \label{table:qatasks}
    \vspace{5pt}
    \resizebox{\textwidth}{!}{
        \begin{tabular}{@{}lllr@{}}
        \toprule
        Question type & Template & Level codename & $\#$ QA pairs \\
        \midrule
        Attribute & What is the color of the $<${\tt shape}$>$? & {\tt color} & $500$ \\
                  & What shape is the $<${\tt color}$>$ object? & {\tt shape} & $500$ \\[0.05in]
        Count     & How many $<${\tt shape}$>$ are there? & {\tt count\_shape} & $200$ \\
                  & How many $<${\tt color}$>$ objects are there? & {\tt count\_color} & $40$ \\[0.05in]
        Exist     & Is there a $<${\tt shape}$>$? & {\tt existence\_shape} & $100$ \\[0.05in]
        Compare $+$ Count
                  & Are there the same number of $<${\tt color1}$>$ objects as $<${\tt color2}$>$ objects?
                        & {\tt compare\_n\_color} & $180$ \\
                  & Are there the same number of $<${\tt shape1}$>$ as $<${\tt shape2}$>$?
                        & {\tt compare\_n\_shape} & $4900$ \\[0.05in]
        Relation $+$ Attribute
                  & What is the color of the $<${\tt shape1}$>$ near the $<${\tt shape2}$>$?
                        & {\tt near\_color} & $24500$ \\
                  & What is the $<${\tt color}$>$ object near the $<${\tt shape}$>$?
                        & {\tt near\_shape} & $25000$ \\[0.05in]
        \bottomrule
        \end{tabular}
    }
    \vspace{-15pt}
\end{table*}

\textbf{Grounded language learning}.
Inspired by the work of~\cite{winograd_cogpsy72} on SHRDLU,
several recent works have explored linguistic representation learning
by grounding language into actions and pixels in physical environments -- in
2D gridworlds~\citep{andreas_icml17,yu_iclr18,misra_acl17},
3D~\citep{chaplot_aaai18,embodiedqa,gordon_cvpr18,cangea_arxiv19,puig_cvpr18,zhu_iccv17,anderson_cvpr18,gupta_cvpr17,zhu_icra17,oh_icml17,shu_iclr18,vogel_acl10, hill2020human}
and textual~\citep{matuszek_iser13,narasimhan_emnlp15} environments.
Closest to our work is the task of Embodied Question Answering~\citep{gordon_cvpr18,embodiedqa,eqa_modular,eqa_multitarget,eqa_matterport} --
where an embodied agent in an environment
(\eg a house) is asked to answer a question (\eg ``What color is the piano?'').
Typical approaches to EmbodiedQA involve training agents to move for the goal of
answering questions. In contrast, our focus is on learning a predictive
model in a \emph{goal-agnostic} exploration phase and using question-answering
as a post-hoc testbed for evaluating the semantic knowledge
that emerges in the agent's representations from predicting the future.

\textbf{Neural population decoding}.
Probing an agent with a QA decoder can be viewed as a variant
of neural population decoding, 
used as an analysis tool in
neuroscience~\citep{georgopoulos_science86,bialek_science91,salinas_jnc94} and
more recently in deep learning~\citep{guo_arxiv18,gregor_neurips19, azar2019world,alain_arxiv16,conneau_acl18,tenney_iclr19}. The idea is to test whether specific information is encoded in a learned representation, by feeding the representation as input to a probe network, generally a classifier trained to extract the desired information. In RL, this is done by training a probe to predict parts of the ground-truth state of the environment, such as an agent's position or orientation, without backpropagating through the agent's internal state.

Prior work has required a separate network to be trained for each probe, even for closely related properties such as position vs. orientation \citep{guo_arxiv18} or grammatical features of different words in the same sentence \citep{conneau_acl18}. Moreover, each probe is designed with property-specific inductive biases, such as convnets for top-down views vs. MLPs for position \citep{gregor_neurips19}. In contrast, we train a single, general-purpose probe network that covers a variety of question types, with an inductive bias for language processing. This generality is possible because of the external conditioning, in the form of the question, supplied to the probe. External conditioning moreover enables agent analysis using novel perturbations of the probe's training questions.

\textbf{Neuroscience}.
Predictive modeling is thought to be a fundamental component of human cognition~\citep{elman_1990,hohwy13,seth15}.
In particular, it has been proposed that perception, learning and decision-making rely on the
minimization of prediction error~\citep{rao_nature99,clark16}.
A well-established strand of work has focused on decoding predictive
representations in brain states~\citep{nortmann_cc13,huth_2016}.
The question of how prediction of sensory experience relates to higher-order conceptual
knowledge is complex and subject to debate~\citep{williams18,roskies_wood_17},
though some have proposed that conceptual knowledge, planning, reasoning, and other
higher-order functions emerge in deeper layers of a predictive network.
We focus on the emergence of propositional knowledge in a predictive agent's internal representations.

\vspace{-5pt}
\section{Environment \& Tasks}

\textbf{Environment.}
We use a Unity-based visually-rich $3$D environment (see Figure \ref{fig:teaser}).
It is a single L-shaped room that can be programmatically
populated with an assortment of objects of different colors at different
spatial locations and orientations. In total, we use a library of $50$ different
objects, referred  to as `shapes' henceforth (\eg chair, teddy, glass, \etc), in $10$ different colors (\eg red, blue, green, \etc).
For a complete list of environment details, see Sec.~\ref{sec:supp_environment}.

At every step, the agent gets a $96 \times 72$ first-person RGB image as its observation,
and the action space consists of movements ({\small {\tt move-}$\{${\tt forward,back,left,right}$\}$}),
turns ({\small {\tt turn-}$\{${\tt up,down,left,right}$\}$}), and object pick-up and manipulation ($4$ DoF:
yaw, pitch, roll, and movement along the axis between the agent and object).
See Table~\ref{table:appendix_action_set} in the Appendix for the full set of actions.

\textbf{Question-Answering Tasks.}
We develop a range of question-answering tasks of varying complexity that
test the agent's local and global scene understanding, visual reasoning, and memory skills.
Inspired by~\cite{johnson_cvpr17,embodiedqa,gordon_cvpr18},
we programmatically generate a dataset of questions
(see Table \ref{table:qatasks}).
These questions ask about the presence or absence of objects ({\small {\tt existence\_shape}}), their attributes ({\small {\tt color, shape}}), counts ({\small {\tt count\_color, count\_shape}}), quantitative comparisons ({\small {\tt compare\_count\_color,
compare\_count\_shape}}), and elementary spatial relations ({\small {\tt near\_color, near\_shape}}).
Unlike the fully-observable setting
in CLEVR~\citep{johnson_cvpr17},
the agent does not get a global view of the environment, and must answer these questions from a sequence of partial egocentric observations.
Moreover, unlike prior work on EmbodiedQA~\citep{gordon_cvpr18,embodiedqa}, the agent is \emph{not} being
trained end-to-end to move to answer questions. It is being trained to explore,
and answers are being decoded (without backpropagating gradients) from its
internal representation. Thus, in order to answer these questions,
the agent \emph{must} learn to encode relevant aspects of the environment in a
representation amenable to easy decoding into symbols (\eg what does the word ``chair''
mean? or what representations does computing ``how many'' require?).

\vspace{-5pt}
\section{Approach}
\label{sec:approach}

\begin{figure*}[t]
    \centering
    \includegraphics[width=0.8\textwidth]{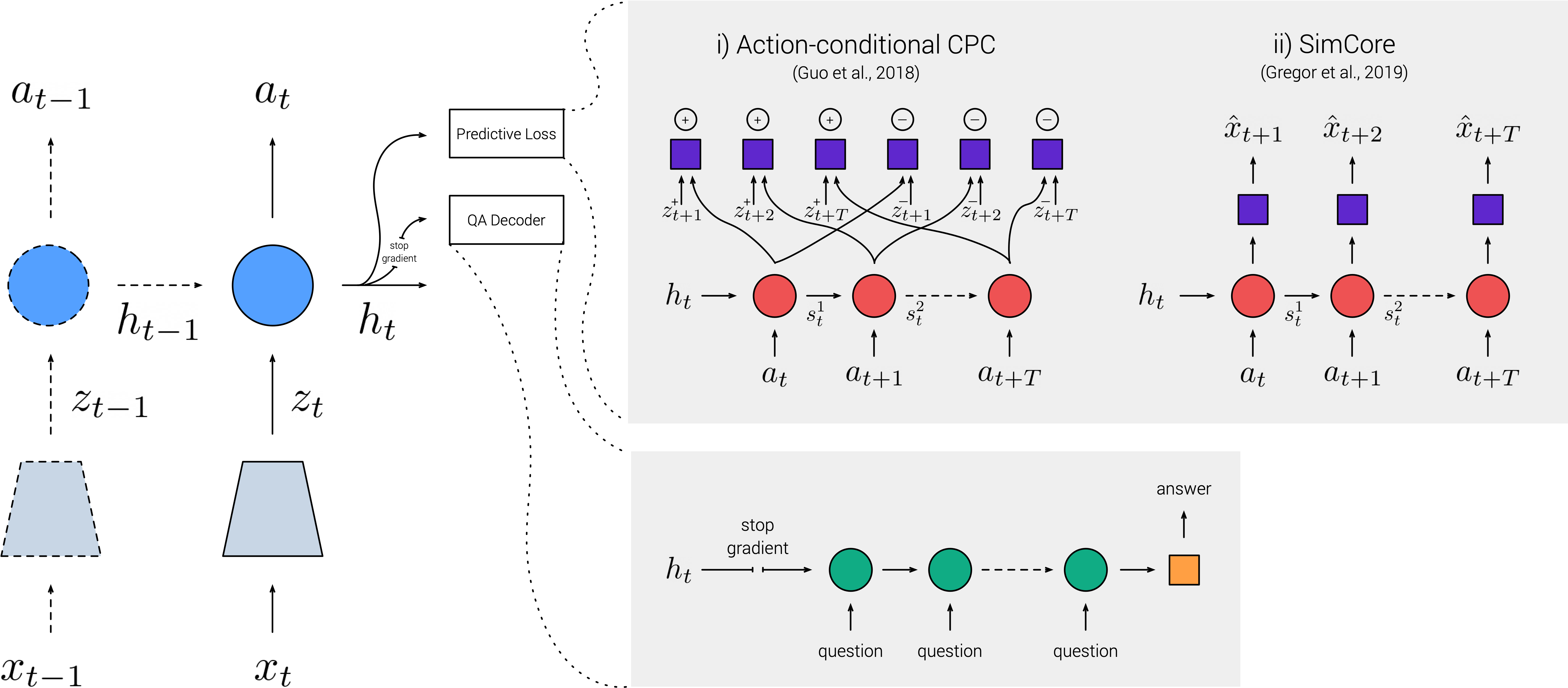}
    \caption{Approach: at every timestep $t$, the agent receives
        an RGB observation $x_t$ as input, processes it using a convolutional neural
        network to produce $z_t$, which is then processed by an LSTM to select action
        $a_t$. The agent learns to explore -- it receives a reward of $1.0$
        for navigating to each new object. As it explores the
        environment, it builds up an internal representation $h_t$, which receives
        pressure from an auxiliary predictive module to capture environment semantics
        so as to accurately predict consequences of its actions multiple steps into the future.
        We experiment with a vanilla LSTM agent and two recent predictive approaches --
        CPC$|$A~\citep{guo_arxiv18} and SimCore~\citep{gregor_neurips19}.
        The internal representations are then probed via a question-answering
        decoder whose gradients are not backpropagated into the agent.
        The QA decoder is an LSTM initialized with $h_t$ and receiving the
        question at every timestep.}
    \label{fig:approach}
    \vspace{-10pt}
\end{figure*}

\textbf{Learning an exploration policy}.
Predictive modeling has proven to be effective for an agent to develop
general knowledge of its environment as it explores and behaves
towards its goal, typically maximising environment returns~\citep{gregor_neurips19,guo_arxiv18}.
Since we wish to evaluate the effectiveness of predictive modeling independent
of the agent's specific goal, we define a simple task that stimulates the agent
to visit all of the `important' places in the environment (\ie to acquire an
exploratory but otherwise task-neutral policy). This is achieved by giving the
agent a reward of $+1.0$ every time it visits an object in the room for the first
time. After visiting all objects, rewards are refreshed and available to be
consumed by the agent again (\ie re-visiting an object the agent has already been to
will now again lead to a $+1.0$ reward), and this process continues for the duration
of each episode ($30$ seconds or $900$ steps).

During training on this exploration task, the agent receives a first-person RGB
observation $x_t$ at every timestep $t$, and processes it using a convolutional neural
network to produce $z_t$. This is input to an LSTM policy whose hidden state is $h_t$
and output a discrete action $a_t$. The agent optimizes the discounted sum of
future rewards using an importance-weighted actor-critic algorithm~\citep{espeholt_arxiv18}.

\textbf{Training the QA-decoder}.
\label{approach:qa}
The question-answering decoder is operationalized as an LSTM that is initialized
with the agent's internal representation $h_t$ and receives the question as input
at every timestep (see Fig.~\ref{fig:approach}). The question is a string that we tokenise
into words and then map to learned embeddings. The question decoder LSTM is
then unrolled for a fixed number of computation steps after which it predicts a
softmax distribution over the vocabulary of one-word answers to questions in Table~\ref{table:qatasks},
and is trained via a cross-entropy loss. Crucially, this QA decoder is trained
independent of the agent policy; \ie gradients from this decoder are not allowed
to flow back into the agent. We evaluate question-answering performance by measuring
top-1 accuracy at the end of the episode --
we consider the agent's top predicted answer at the last time step of the episode and compare that with the ground-truth answer.

The QA decoder can be seen as a general purpose decoder trained to extract
object-specific knowledge from the agent's internal state without affecting the agent itself.
If this knowledge is not retained in the agent's internal state,
then this decoder will not be able to extract it. This is
an important difference with respect to prior work~\citep{gordon_cvpr18,embodiedqa}
-- wherein agents were trained to move to answer questions, \ie
all parameters had access to linguistic information.
Recall that the agent's navigation policy has been trained for exploration,
and so the visual information required to answer a question need not be present in the
observation at the end of the episode. Thus, through question-answering,
we are evaluating the degree to which agents encode relevant aspects of the environment
(object colors, shapes, counts, spatial relations)
in their internal representations
\emph{and} maintain this information in memory
beyond the point at which it was initially received.
See~\ref{appendix_qa_network} for more details about the QA decoder.

\subsection{Auxiliary Predictive Losses}
We augment the baseline architecture described above with an auxiliary predictive
head consisting of a simulation network (operationalized
as an LSTM) that is initialized with the agent's internal state $h_t$ and deterministically
simulates future latent states $s^1_t, \ldots, s^{k}_t, \ldots$ in an open-loop manner,
receiving the agent's action sequence as input. We evaluate two predictive losses --
action-conditional CPC~\citep{guo_arxiv18} and SimCore~\citep{gregor_neurips19}.
See Fig.~\ref{fig:approach} for overview, \ref{appendix_agents} for details.

\textbf{Action-conditional CPC}~(CPC$|$A, \cite{guo_arxiv18})
makes use of a noise contrastive estimation model to
discriminate between true observations processed by the convolutional neural network
$z^+_{t+k}$ ($k$ steps into the future) and negatives randomly sampled from
the dataset~$z^-_{t+k}$, in our case from other episodes in the minibatch.
Specifically, at each timestep $t+k$ (up to a maximum),
the output of the simulation core $s^{k}_t$ and $z^+_{t+k}$
are fed to an MLP to predict $1$, and $s^{k}_t$ and $z^-_{t+k}$ are used to predict $0$.

\textbf{SimCore}~\citep{gregor_neurips19} uses the simulated state $s_t^k$ to condition a generative model
based on ConvDRAW~\citep{gregor_neurips16} and GECO~\citep{rezende_arxiv18}
that predicts the distribution of true observations
$p(x_{t+k}|h_t, a_{t,...,(t+k)})$ in pixel space.

\begin{figure*}[t]
    \centering
    \includegraphics[width=0.8\textwidth]{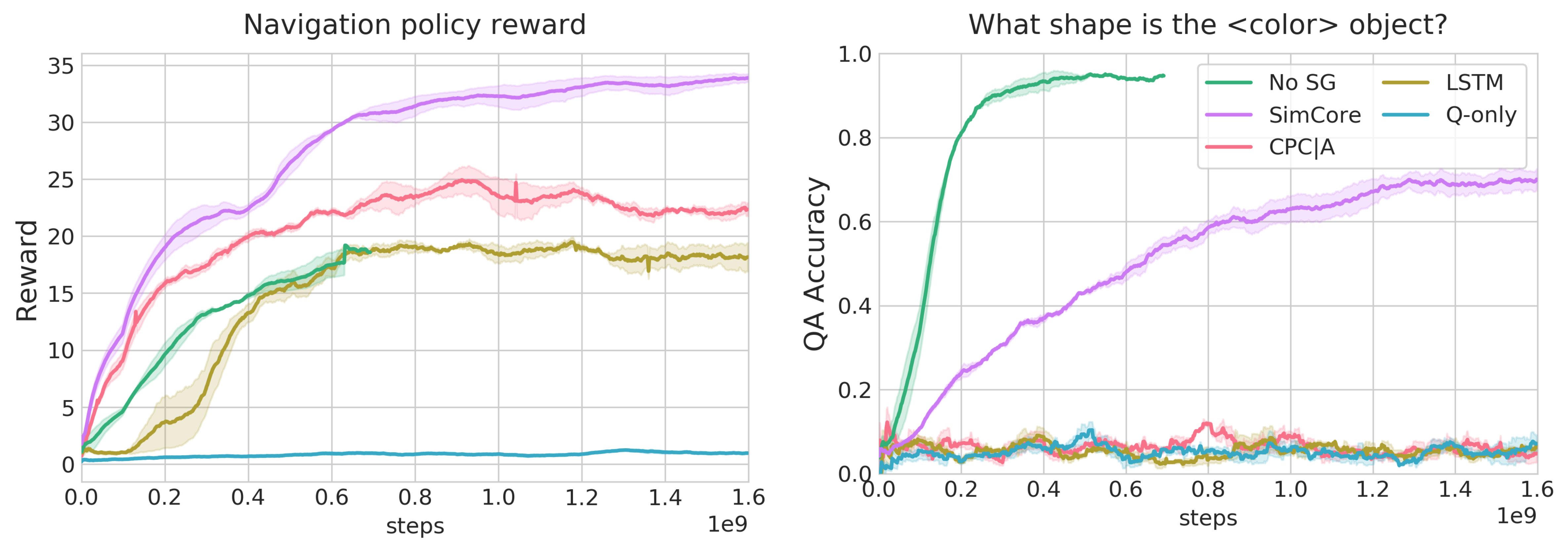}
    \vspace{-10pt}
    \caption{{\small L -- Reward in an episode. R -- Top-$1$ QA accuracy.
        Averaged over $3$ seeds. Shaded region is $1$ SD.}}
    \label{fig:fig_3}
    \vspace{-10pt}
\end{figure*}

\textbf{Baselines}.
We evaluate and compare the above approaches
with 1) a vanilla RL agent without any auxiliary predictive losses
(referred to as `LSTM'), and 2) a question-only agent that receives
zero-masked observations as input and is useful to measure biases in our question-answering
testbed. Such a baseline is critical, particularly when working with simulated environments,
as it can uncover biases in the environment's generation of tasks that can result in strong
but uninteresting performance from agents capable of powerful function approximation~\citep{thomason_naacl19}.

\textbf{No stop gradient}.
We also compare against an agent without blocking the QA decoder gradients
(labeled `No SG'). This model differs from the above in that
it is trained end-to-end -- with supervision -- to answer the set of questions
in addition to the exploration task. Hence, it represents an agent receiving
privileged information about how to answer and its performance
provides an upper bound for how challenging these question-answering tasks
are in this context.

\begin{table*}[t]
    \vspace{-10pt}
    \caption{Top-$1$ accuracy on question-answering tasks.}
    \vspace{-15pt}
    \label{table:results_v1}
        \centering
        \resizebox{\textwidth}{!}{
        \begin{tabular}{lclclclclclclclclclc}
        &  \mcrot{1}{l}{45}{Overall}
        && \mcrot{1}{l}{45}{\tt{shape}}
        && \mcrot{1}{l}{45}{\tt{color}}
        && \mcrot{1}{l}{45}{\tt{exist}}
        && \mcrot{1}{l}{45}{\tt{count\_shape}}
        && \mcrot{1}{l}{45}{\tt{count\_color}}
        && \mcrot{1}{l}{45}{\tt{compare\_n\_color}}
        && \mcrot{1}{l}{45}{\tt{compare\_n\_shape}}
        && \mcrot{1}{l}{45}{\tt{near\_shape}}
        && \mcrot{1}{l}{45}{\tt{near\_color}} \\
        \midrule
        Baseline: Question-only &
            $29$ {\scriptsize $\pm~3$} &&
            $04$ {\scriptsize $\pm~2$} &&
            $10$ {\scriptsize $\pm~2$} &&
            $63$ {\scriptsize $\pm~4$} &&
            $24$ {\scriptsize $\pm~3$} &&
            $24$ {\scriptsize $\pm~3$} &&
            $49$ {\scriptsize $\pm~3$} &&
            $70$ {\scriptsize $\pm~3$} &&
            $04$ {\scriptsize $\pm~2$} &&
            $09$ {\scriptsize $\pm~3$} \\
        \midrule
        LSTM &
            $31$ {\scriptsize $\pm~4$} &&
            $04$ {\scriptsize $\pm~1$} &&
            $10$ {\scriptsize $\pm~2$} &&
            $54$ {\scriptsize $\pm~6$} &&
            $34$ {\scriptsize $\pm~3$} &&
            $38$ {\scriptsize $\pm~3$} &&
            $53$ {\scriptsize $\pm~3$} &&
            $70$ {\scriptsize $\pm~3$} &&
            $04$ {\scriptsize $\pm~2$} &&
            $09$ {\scriptsize $\pm~3$} \\
        CPC$\vert$A &
            $32$ {\scriptsize $\pm~3$} &&
            $06$ {\scriptsize $\pm~2$} &&
            $08$ {\scriptsize $\pm~2$} &&
            $64$ {\scriptsize $\pm~3$} &&
            $\mathbf{39}$ {\scriptsize $\pm~3$} &&
            $39$ {\scriptsize $\pm~3$} &&
            $50$ {\scriptsize $\pm~4$} &&
            $70$ {\scriptsize $\pm~3$} &&
            $06$ {\scriptsize $\pm~2$} &&
            $10$ {\scriptsize $\pm~3$} \\
        SimCore &
            $\mathbf{60}$ {\scriptsize $\pm~3$} &&
            $\mathbf{72}$ {\scriptsize $\pm~3$} &&
            $\mathbf{81}$ {\scriptsize $\pm~3$} &&
            $\mathbf{72}$ {\scriptsize $\pm~3$} &&
            $\mathbf{39}$ {\scriptsize $\pm~3$} &&
            $\mathbf{57}$ {\scriptsize $\pm~3$} &&
            $\mathbf{56}$ {\scriptsize $\pm~3$} &&
            $\mathbf{73}$ {\scriptsize $\pm~3$} &&
            $\mathbf{30}$ {\scriptsize $\pm~3$} &&
            $\mathbf{59}$ {\scriptsize $\pm~3$} \\
        \midrule
        Oracle: No SG &
            $63$ {\scriptsize $\pm~3$} &&
            $96$ {\scriptsize $\pm~2$} &&
            $81$ {\scriptsize $\pm~2$} &&
            $60$ {\scriptsize $\pm~3$} &&
            $45$ {\scriptsize $\pm~3$} &&
            $57$ {\scriptsize $\pm~3$} &&
            $51$ {\scriptsize $\pm~3$} &&
            $76$ {\scriptsize $\pm~3$} &&
            $41$ {\scriptsize $\pm~3$} &&
            $72$ {\scriptsize $\pm~3$} \\
        \bottomrule
        \end{tabular}
        \vspace{-10pt}
        }
    \vspace{5pt}
    \vspace{-10pt}
\end{table*}

\vspace{-5pt}
\section{Experiments \& Results}

\subsection{Question-Answering Performance}
We begin by analyzing performance on a single question --
    {\small {\tt shape}} -- which are of the form
    ``what shape is the $<$color$>$ object?''.
    Figure~\ref{fig:fig_3} shows the average reward accumulated by the agent in one
    episode (left) and the QA accuracy at the last timestep of the episode (right)
    for all approaches over the course of training.
    We make the following observations:

\begin{figure*}[h]
    \centering
    \begin{subfigure}[b]{0.62\textwidth}
        \centering
        \includegraphics[width=\textwidth]{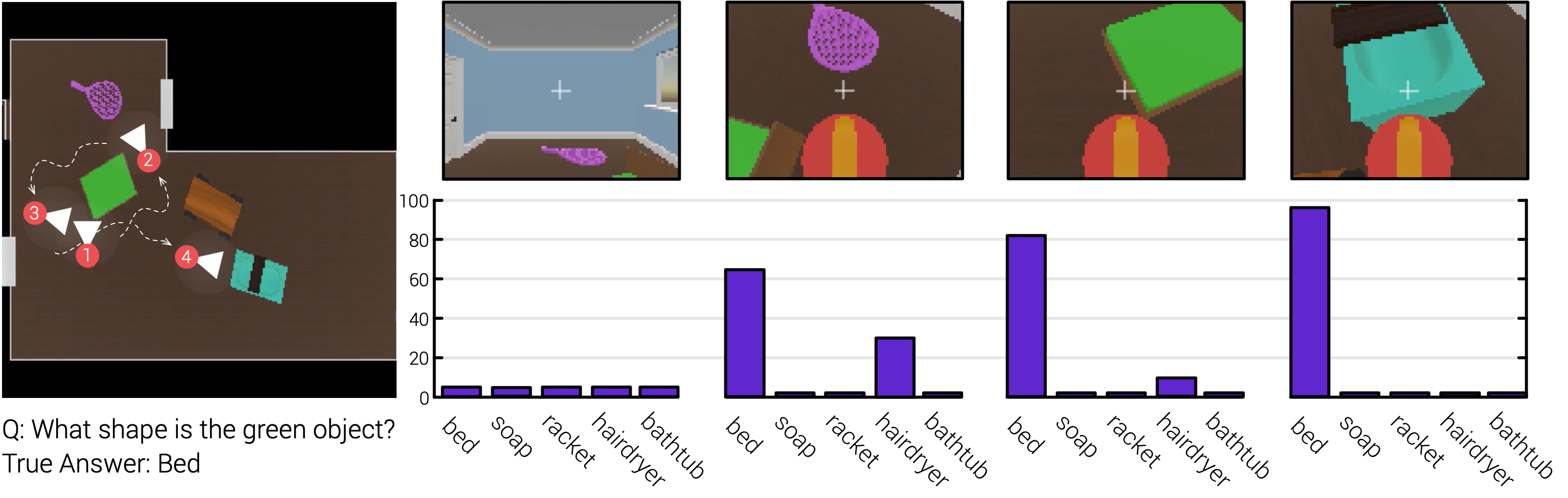}
    \end{subfigure}
    \begin{subfigure}[b]{0.37\textwidth}
        \centering
        \includegraphics[width=\textwidth]{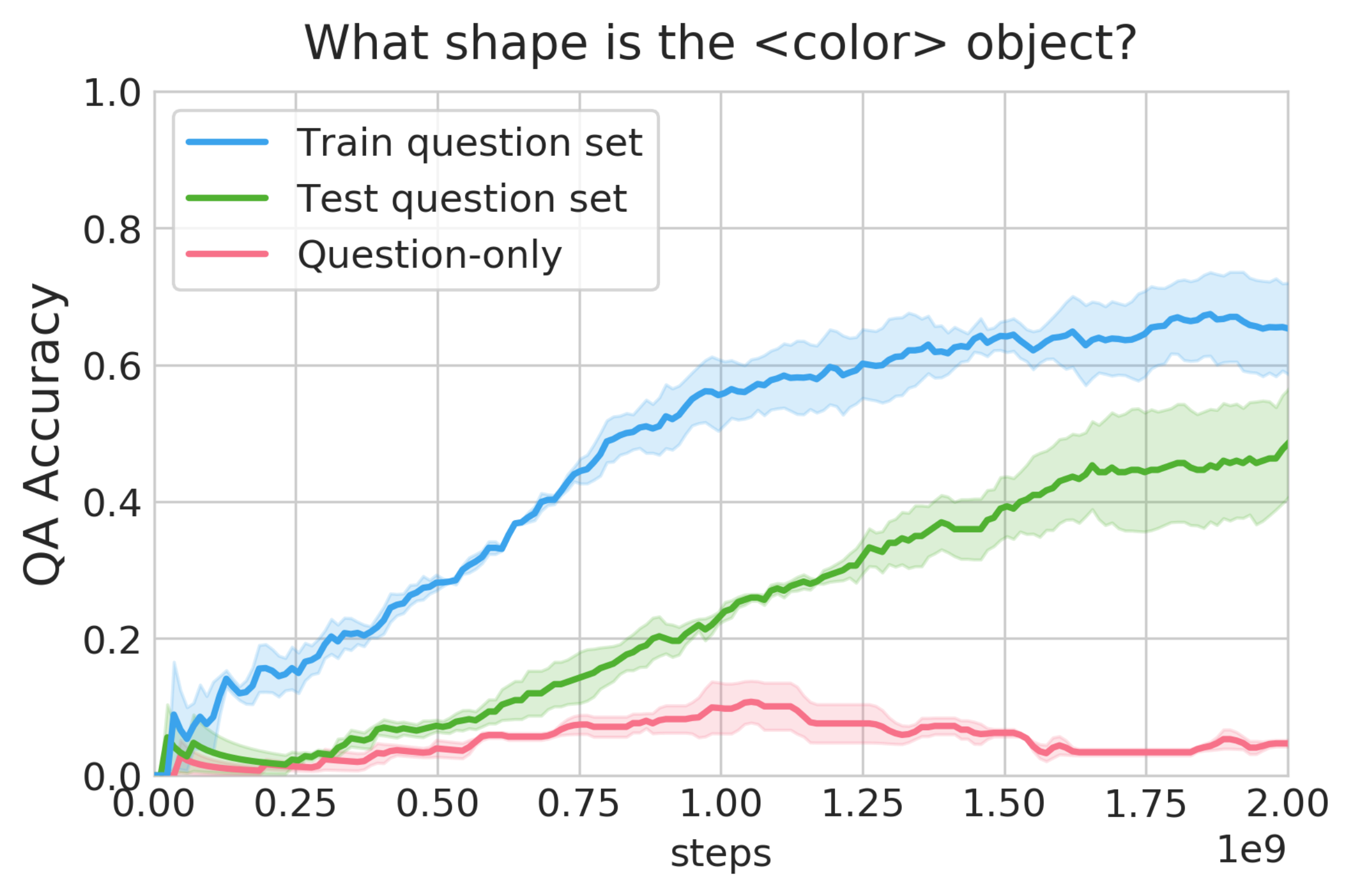}

    \end{subfigure}
    \vspace{-15pt}
    \caption{(Left): Sample trajectory ($1 \rightarrow 4$) and QA
            decoding predictions (for top 5 most probable answers) for the
            `What shape is the green object?' from SimCore.
            Note that top-down map is not available to the agent.
            (Right): QA accuracy on disjoint train and test splits.}
    \vspace{-10pt}
    \label{fig:fig_4}
\end{figure*}

\vspace{-10pt}

\begin{itemize}
\itemsep1pt
\item \textbf{All agents learn to explore}. With the exception `question-only', all agents achieve high reward on the exploration task.
    This means that they visited all objects in the room more than once each and therefore, in principle,
    have been exposed to sufficient information to answer all questions.
\item \textbf{Predictive models aid navigation}. Agents equipped with auxiliary predictive losses -- CPC$|$A and SimCore --
    collect the most rewards, suggesting that predictive modeling helps navigate the environment efficiently.
    This is consistent with findings in~\cite{gregor_neurips19}.
\item \textbf{QA decoding from LSTM and CPC$|$A representations is no better than chance}.
\item \textbf{SimCore's representations lead to best QA accuracy}.
    SimCore gets to a QA accuracy of ${\sim}72\%$ indicating
    that its representations best capture propositional knowledge
    and are best suited for decoding answers to questions.
    Figure~\ref{fig:fig_4} (Left) shows example predictions.
\item \textbf{Wide gap between SimCore and No SG}.
    There is a ${\sim}24\%$ gap between
    SimCore and the No SG oracle, suggesting scope for better
    auxiliary predictive losses.
\end{itemize}
\vspace{-5pt}

It is worth emphasizing that answering this {\tt shape} question from observations
is not a challenging task in and of itself. The No SG agent, which is trained
end-to-end to optimize both for exploration and QA, achieves almost-perfect accuracy (${\sim}96\%$).
The challenge arises from the fact that we are not training the agent end-to-end
-- from pixels to navigation to QA -- but decoding the answer from the agent's internal state,
which is learned agnostic to the question.
The answer can only be decoded if the agent's internal state contains
relevant information represented in an easily-decodable way.

\begin{figure*}[t]
    \centering
    \begin{subfigure}{0.55\textwidth}
        \centering
        \includegraphics[width=\textwidth]{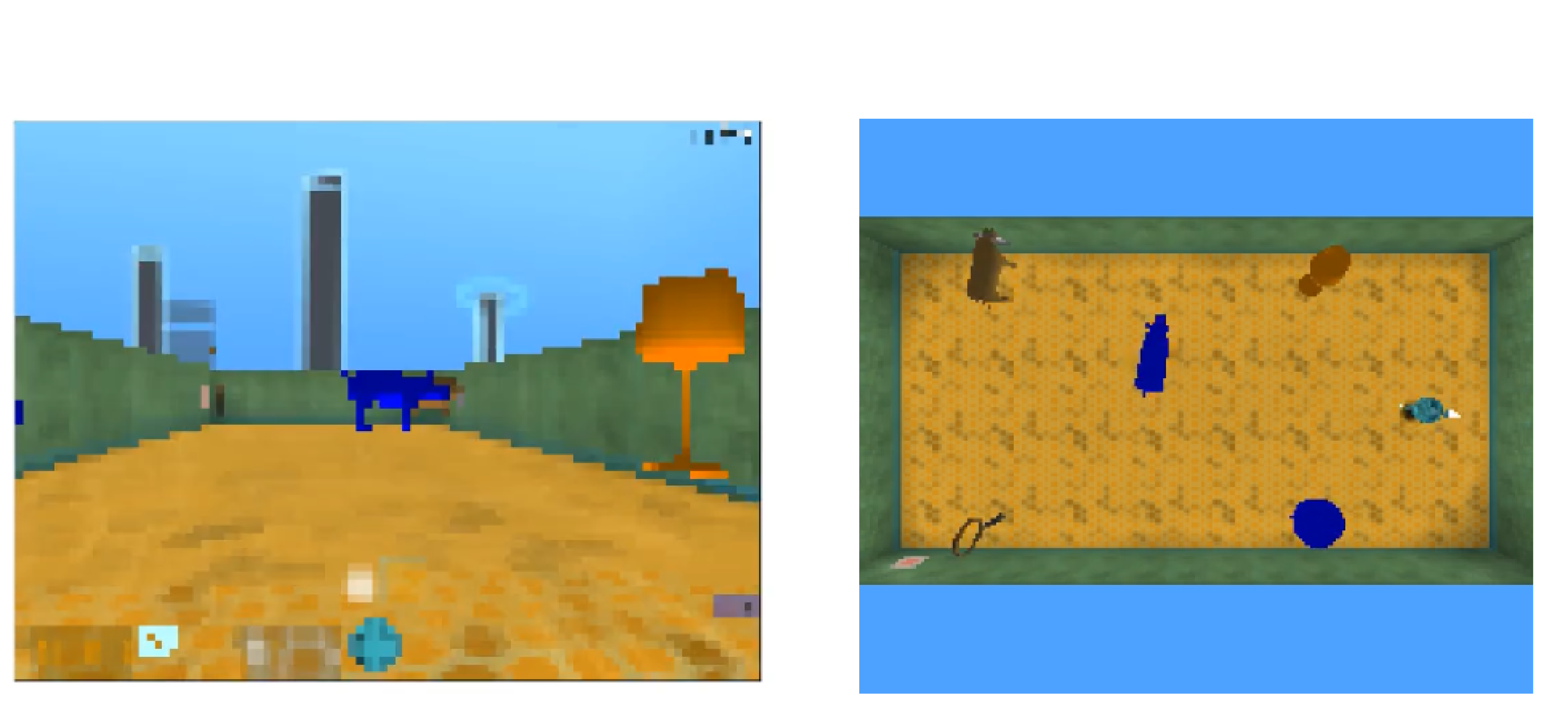}
        \vspace{5pt}
    \end{subfigure}
    \quad
    \begin{subfigure}{0.4\textwidth}
        \centering
        \includegraphics[width=\textwidth]{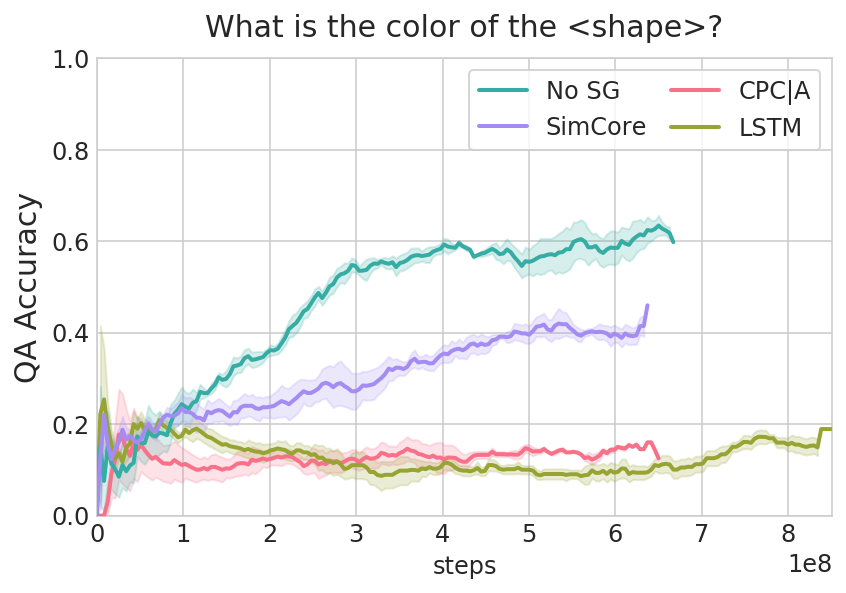}
        \vspace{5pt}
    \end{subfigure}
    \vspace{-30pt}
    \caption{(Left) DeepMind Lab environment~\citep{Beattie16_deepmind_lab}:
        Rectangular-shaped room with 6 randomly selected objects out of a
        pool of 20 different objects of different colors. (Right) QA accuracy for {\tt color} questions (What is the color of the $<${\tt shape}$>$?) in DeepMind Lab. Consistent with results in the main paper, internal representations of the SimCore agent lead to the highest accuracy while CPC$|$A and LSTM perform worse and similar to each other.}
    \label{fig:fig_dmlab}
    \vspace{-15pt}
\end{figure*}

\textbf{Decoder complexity}.
To explore the possibility that answer-relevant information is present in the
agent's internal state but requires a more powerful decoder, we experiment with
QA decoders of a range of depths. As detailed in Figure~\ref{fig:sup_fig_3a} in
the appendix, we find that using a deeper QA decoder with SimCore does lead to
higher QA accuracy (from $1 \rightarrow 12$ layers), although greater decoder
depths become detrimental after $12$ layers. Crucially, however, in the
non-predictive LSTM agent, the correct answer cannot be decoded
irrespective of QA decoder capacity.
This highlights an important aspect of our question-answering evaluation
paradigm -- that while the absolute accuracy at answering questions may also depend
on decoder capacity, relative differences provide an informative comparison
between internal representations developed by different agents.

Table~\ref{table:results_v1} shows QA accuracy for all QA tasks
(see Figure \ref{fig:sup_fig_3b} in appendix for training curves).
The results reveal large variability in difficulty across question types.
Questions about attributes ({\tt color} and {\tt shape}), which can be answered
from a single well-chosen frame of visual experience, are the easiest,
followed by spatial relationship questions ({\tt near\_color} and {\tt near\_shape}),
and the hardest are counting questions ({\tt count\_color} and {\tt count\_shape}). We further note that:

\vspace{-10pt}

\begin{itemize}
\itemsep1pt
    \item \textbf{All agents perform better than the question-only baseline}, which captures
        any biases in the environment or question distributions
        (enabling strategies such as constant prediction of the most-common answer).
    \item \textbf{CPC$\vert$A representations are not better than LSTM on most question types}.
    \item \textbf{SimCore representations achieve higher QA accuracy than other approaches},
        substantially above the question-only baseline on {\small {\tt count\_color}}
        ($57\%$ \vs $24\%$), {\small {\tt near\_shape}} ($30\%$ \vs $4\%$) and
        {\small {\tt near\_color}} ($59\%$ \vs $9\%$), demonstrating a strong tendency
        for encoding and retaining information about object identities,
        properties, and both spatial and temporal relations.
\end{itemize}

Finally, as before, the No SG agent trained to answer questions without stopped
gradients achieves highest accuracy for most questions, although not all -- perhaps
due to trade-offs between simultaneously optimizing performance for different
QA losses and the exploration task.

\subsection{Compositional Generalization}

While there is a high degree of procedural randomization in our environment and QA tasks,
overparameterized neural-network-based models in limited environments are always prone to overfitting or rote memorization.
We therefore constructed a test of the generality of the information encoded in the internal state of an agent.
The test involves a variant of the {\tt shape} question type (\ie questions like ``what shape is the $<${\tt color}$>$ object?''),
but in which the possible question-answer pairs are partitioned into mutually exclusive training and test splits.
Specifically, the test questions are constrained such that they are compositionally
novel -- the {\small $<${\tt color}, {\tt shape}$>$} combination involved in the
question-answer pair is never observed during training, but both attributes are observed in other contexts.
For instance, a test question-answer pair ``Q: what shape is the \textbf{blue} object?, A: \textbf{table}''
is excluded from the training set of the QA decoder, but ``Q: what shape is the \textbf{blue} object?, A: \textbf{car}''
and ``Q: What shape is the \textbf{green} object?, A: \textbf{table}'' are part of the training set (but not the test set).

We evaluate the SimCore agent on this test of generalization (since other agents perform poorly
on the original task). Figure \ref{fig:fig_4} (right) shows that
the QA decoder applied to SimCore's internal states performs at substantially above-chance (and all baselines) on the held-out test questions (although somewhat lower than training performance). This indicates that the QA decoder extracts and applies information in a comparatively factorized (or compositional) manner, and suggests (circumstantially) that the knowledge acquired by the SimCore agent may also be represented in this way.

\subsection{Robustness of the results}

To check if our results are robust to the choice of environment, we
developed a similar setup using the DeepMind Lab environment ~\citep{Beattie16_deepmind_lab} and ran the same experiments \emph{without} any change in hyperparameters.

The environment consists of a rectangular room that is populated with a random selection of objects of different shapes and colors in each episode. There are 6 distinct objects in each room, selected from a pool of 20 objects and 9 different colors. We use a similar exploration reward structure as in our earlier environment to train the agents to navigate and observe all objects. Finally, in each episode, we introduce a question of the form
`What is the color of the $<${\tt shape}$>$?' where $<${\tt shape}$>$ is
replaced by the name of an object present in the room.

Figure \ref{fig:fig_dmlab} shows question-answering accuracies in
the DeepMind Lab environment. Consistent with the results presented above, internal representations of the SimCore agent lead to the highest
answering accuracy while CPC$|$A and the vanilla LSTM agent perform worse and
similar to each other. Crucially, for running experiments in DeepMind Lab, we
\emph{did not} change any hyperparameters from the experimental setup described before. This demonstrates that our approach is not specific to a
single environment and that it can be readily applied in a variety of settings.

\section{Discussion}

Developing agents with world models of their environments is an important problem in AI. To do so, we need tools to evaluate and diagnose the internal representations forming these world models in addition to studying task performance. Here, we marry together population or glass-box decoding techniques with a question-answering paradigm to discover how much propositional (or declarative) knowledge agents acquire as they explore their environment. 

We started by developing a range of question-answering tasks in a visually-rich $3$D environment, serving as a diagnostic test of an agent's scene understanding, visual reasoning, and memory skills. Next, we trained agents to optimize an exploration objective with and without auxiliary self-supervised predictive losses, and evaluated the representations they form as they explore an environment, via this question-answering testbed. We compared model-free RL agents alongside agents that make egocentric visual predictions and found that the latter (in particular SimCore~\citep{gregor_neurips19}) are able to reliably capture detailed propositional knowledge in their internal states, which can be decoded as answers to questions, while non-predictive agents do not, even if they optimize the exploration objective well.

Interestingly, not all predictive agents are equally good at acquiring knowledge of objects, relations and quantities. We compared a model learning the probability distribution of future frames in pixel space via a generative model (SimCore~\citep{gregor_neurips19}) with a model based on discriminating frames through contrastive estimation (CPC$|$A~\citep{guo_arxiv18}). We found that while both learned to navigate well, only the former developed representations that could be used for answering questions about the environment.
\cite{gregor_neurips19} previously showed that the choice of predictive model has a significant impact on the ability to decode an agent's position and top-down map reconstructions of the environment from its internal representations. Our experiments extend this result to decoding factual knowledge, and demonstrate that the question-answering approach has utility for comparing agents.

Finally, the fact that we can even decode answers to questions from an agent's internal representations learned solely from egocentric future predictions, without exposing the agent itself directly to knowledge in propositional form, is encouraging. It indicates that the agent is learning to form and maintain invariant object identities and properties (modulo limitations in decoder capacity) in its internal state \emph{without explicit supervision}.

It is ${\sim}$30 years since~\cite{elman_1990} showed how syntactic structures and
semantic organization can emerge in the units of a neural network as a consequence of the simple objective of predicting the next word in a sequence. This work corroborates Elman's findings, showing that language-relevant general knowledge can emerge in a situated neural-network agent that predicts future low-level visual observations via sufficiently powerful generative mechanism. The result also aligns with perspectives that emphasize the importance of between sensory modalities in supporting the development of conceptual or linguistic knowledge~\cite{mcclell2019extending}. Our study is a small example of how language can be used as a channel to probe and understand what exactly agents can learn from their environments. We hope it motivates future research in evaluating predictive agents using natural linguistic interactions.


\bibliographystyle{icml2020}
\bibliography{main}

\clearpage

\appendix
\section{Appendix}

\subsection{Network architectures and Training setup} \label{appendix_network_training}

\subsubsection{Importance Weighted Actor-Learner Architecture}

Agents were trained using the IMPALA framework~\citep{espeholt_arxiv18}. Briefly, there are N parallel `actors' collecting experience from the environment in a replay buffer and one learner taking batches of trajectories and performing the learning updates. During one learning update the agent network is unrolled, all the losses (RL and auxiliary ones) are evaluated and the gradients computed.

\subsubsection{Agents} \label{appendix_agents}

\textbf{Input encoder}
To process the frame input, all models in this work use a residual network \citep{he_cvpr16} of 6 64-channel ResNet blocks with rectified linear activation functions and bottleneck channel of size 32. We use strides of (2, 1, 2, 1, 2, 1) and don't use batch-norm. Following the convnet we flatten the ouput and use a linear layer to reduce the size to 500 dimensions. Finally, we concatenate this encoding of the frame together with a one hot encoding of the previous action and the previous reward.

\textbf{Core architecture} The recurrent core of all agents is a 2-layer LSTM with 256 hidden units per layer. At each time step this core consumes the input embedding described above and updates its state. We then use a 200 units single layer MLP to compute a value baseline and an equivalent network to compute action logits, from where one discrete action is sampled.

\textbf{Simulation Network} Both predictive agents have a simulation network with the same architecture as the agent's core. This network is initialized with the agent state at some random time $t$ from the trajectory and unrolled forward for a random number of steps up to 16, receiving only the actions of the agent as inputs. We then use the resulting LSTM hidden state as conditional input for the prediction loss (SimCore or CPC$\vert$A).

\textbf{SimCore} We use the same architecture and hyperparameters described in \cite{gregor_neurips19}. The output of the simulation network is used to condition a Convolutional DRAW~\citep{gregor_neurips16}. This is a conditional deep variational auto-encoder with recurrent encoder and decoder using convolutional operations and a canvas that accumulates the results at each step to compute the distribution over inputs. It features a recurrent prior network that receives the conditioning vector and computes a prior over the latent variables. See more details in \cite{gregor_neurips19}.

\textbf{Action-conditional CPC} We replicate the architecture used in \cite{guo_arxiv18}. CPC$\vert$A uses the output of the simulation network as input to an MLP that is trained to discriminate true versus false future frame embedding. Specifically, the simulation network outputs a conditioning vector after $k$ simulation steps which is concatenated with the frame embedding $z_{t+k}$ produced by the image encoder on the frame $x_{t+k}$ and sent through the MLP discriminator. The discriminator has one hidden layer of 512 units, ReLU activations and a linear output of size 1 which is trained to binary classify true embeddings into one class and false embeddings into another. We take the negative examples from random time points in the same batch of trajectories.

\subsubsection{QA network architecture} \label{appendix_qa_network}

\textbf{Question encoding}
The question string is first tokenized to words and then mapped to integers corresponding to vocabulary indices. These are then used to lookup 32-dimensional embeddings for each word. We then unroll a 64-units single-layer
LSTM for a fixed number of 15 steps. The language representation is then
computed by summing the hidden states for all time steps.

\textbf{QA decoder}. To decode answers from the internal state of the agents we use a second LSTM initialized with the internal state of the agent's LSTM and unroll it for a fixed number of steps, consuming the question embedding at each step. The results reported in the main section were computed using 12 decoding steps. The terminal state is sent through a two-layer MLP (sizes 256, 256) to compute a vector of answer logits with the size of the vocabulary and output the top-1 answer.

\subsubsection{Hyper-parameters}

The hyper-parameter values used in all the experiments are in Table \ref{tab:simcore_hyperparams}.

\begin{table*}[h]
\centering
    \begin{tabular}{|ll|}
        \hline
        \textbf{Agent} & \\ \hline
        Learning rate           &  1e-4  \\ \hline
        Unroll length           & 50  \\ \hline
        Adam $\beta_1$          & 0.90  \\ \hline
        Adam $\beta_2$          & 0.95  \\ \hline
        Policy entropy regularization &  0.0003  \\ \hline
        Discount factor & 0.99  \\ \hline
        No. of ResNet blocks & 6  \\ \hline
        No. of channel in ResNet block & 64  \\ \hline
        Frame embedding size & 500  \\ \hline
        No. of LSTM layers & 2  \\ \hline
        No. of units per LSTM layer & 256  \\ \hline
        No. of units in value MLP & 200  \\ \hline
        No. of units in policy MLP & 200  \\ \hline
        \hline
        \textbf{Simulation Network}  & \\ \hline
        Overshoot length        & 16  \\ \hline
        No. of LSTM layers & 2  \\ \hline
        No. of units per LSTM layer & 256  \\ \hline
        No. of simulations per trajectory & 6  \\ \hline
        No. of evaluations per overshoot  & 2  \\ \hline
        \hline
        \textbf{SimCore}  & \\ \hline
        No. of ConvDRAW Steps & 8\\ \hline
        GECO kappa & 0.0015 \\ \hline
                \hline
        \textbf{CPC$\vert$A}  & \\ \hline
        MLP discriminator size        & 64  \\ \hline
        \hline
        \textbf{QA network}  & \\ \hline
        Vocabulary size        & 1000  \\ \hline
        Maximum question length & 15  \\ \hline
        No. of units in Text LSTM encoder & 64  \\ \hline
        Question embedding size & 32  \\ \hline
        No. of LSTM layers in question decoder & 2  \\ \hline
        No. of units per LSTM layer & 256  \\ \hline
        No. of units in question decoder MLP & 200  \\ \hline
        No. of decoding steps & 12  \\ \hline
        \end{tabular}
        \vspace{5pt}
    \caption{Hyperparameters.}
    \label{tab:simcore_hyperparams}
\end{table*}

\subsubsection{Negative sampling strategies for CPC$\vert$A}

\begin{figure*}[t]
    \centering
    \begin{subfigure}{0.45\textwidth}
        \centering
        \includegraphics[width=\textwidth]{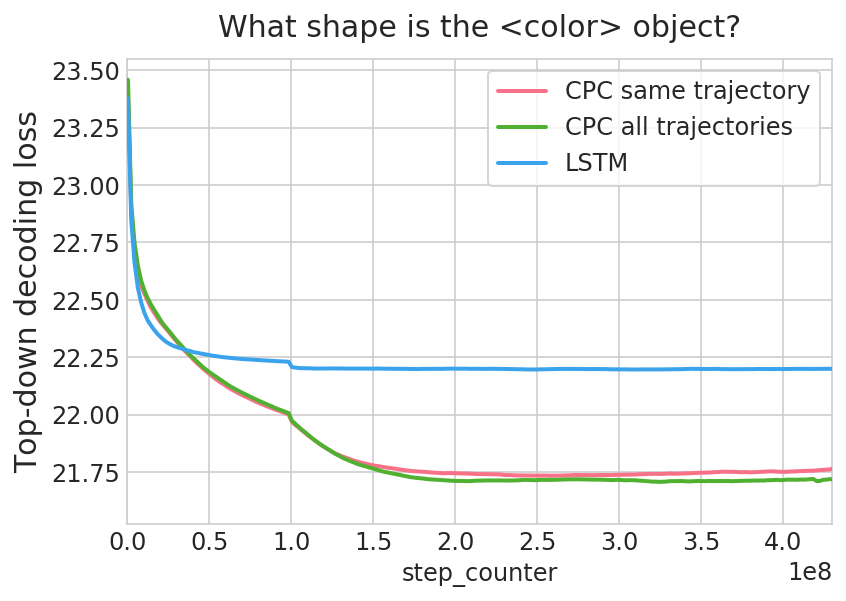}
        \caption{{\scriptsize To test whether the CPC$\vert$A loss provided improved representations we reconstructed the environment top-down view, similar to \cite{gregor_neurips19}. Indeed the reconstruction loss is lower for CPC$\vert$A than for the LSTM agent.}}
        \label{fig:fig_sup_cpc_1}
    \end{subfigure}
    \begin{subfigure}{0.45\textwidth}
        \centering
        \includegraphics[width=\textwidth]{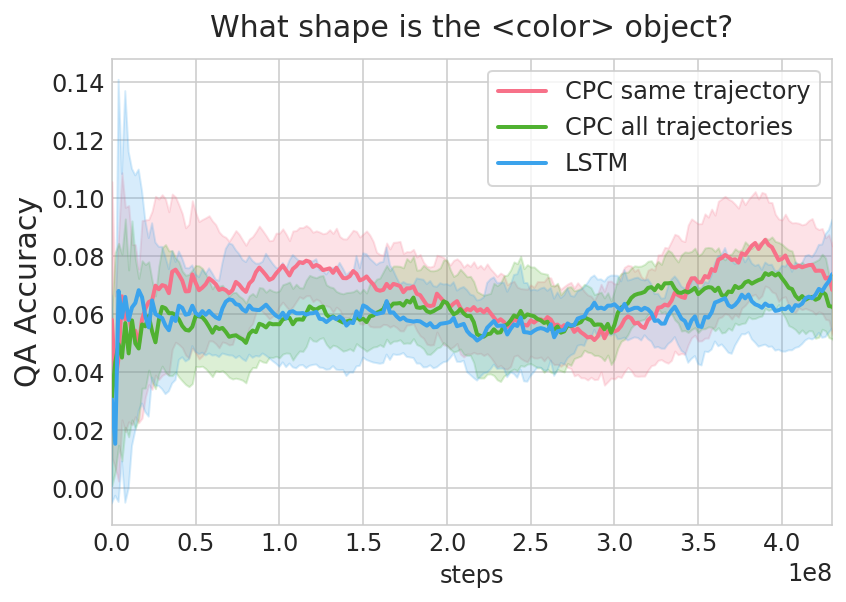}
        \caption{{\scriptsize QA accuracy for the CPC$\vert$A agent is not better than the LSTM agent, for both sampling strategies of negatives.}}
        \label{fig:fig_sup_cpc_2}
    \end{subfigure}
    \vspace{5pt}
    \caption{}
    \vspace{-10pt}
\end{figure*}

We experimented with multiple sampling strategies for the CPC$\vert$A agent (whether or not negative examples are sampled from the same trajectory, the number of contrastive prediction steps, the number of negative examples). We report the best results in the main text. The CPC$\vert$A agent did provide better representations of the environment than the LSTM-based agent, as shown by the top-down view reconstruction loss (Figure \ref{fig:fig_sup_cpc_1}). However, none of the CPC$\vert$A agent variations that we tried led to better-than-chance question-answering accuracy. As an example, in Figure \ref{fig:fig_sup_cpc_2} we compare sampling negatives from the same trajectory or from any trajectory in the training batch.

\subsection{Effect of QA network depth}

\begin{figure*}[h]
    \centering
    \includegraphics[width=\textwidth]{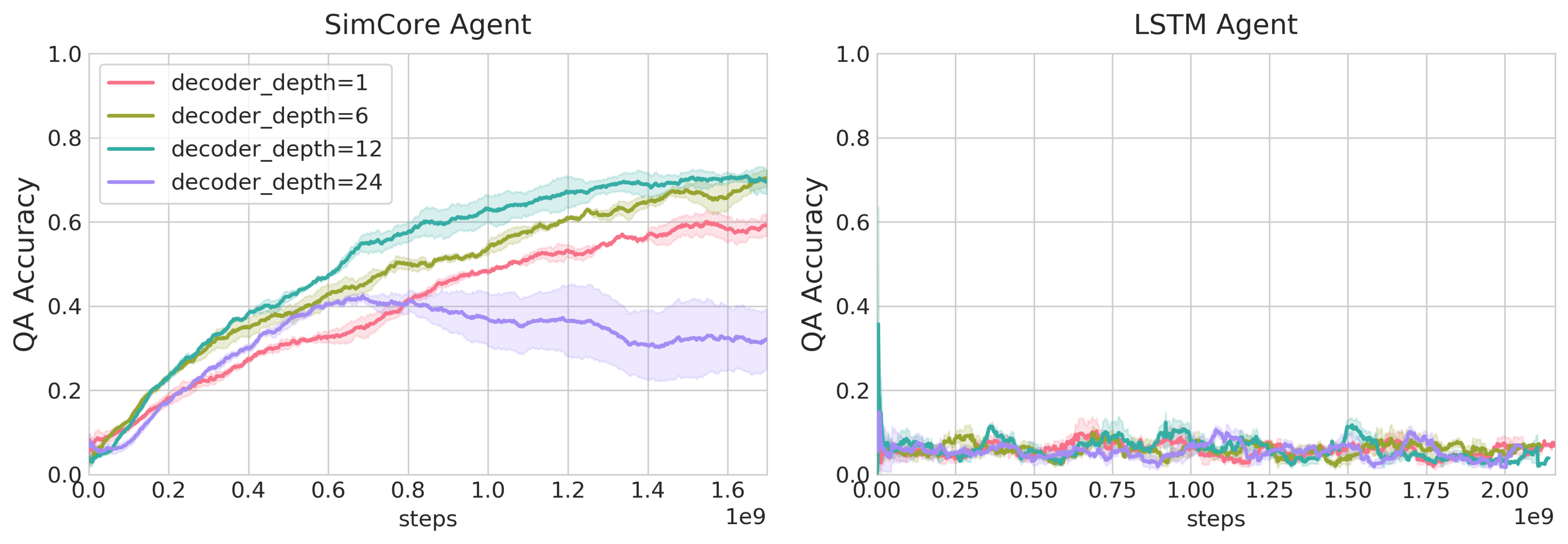}
    \caption{Answer accuracy over training for increasing QA decoder's depths. Left subplot shows the results for the SimCore agent and right subplot for the LSTM baseline. For SimCore, the QA accuracy increases with the decoder depth, up to 12 layers. For the LSTM agent, QA accuracy is not better than chance regardless of the capacity of the QA network.}
    \label{fig:sup_fig_3a}
\end{figure*}

To study the effect of the QA network capacity on the answer accuracy, we tested decoders of different depths applied to both the SimCore and the LSTM agent's internal representations (\ref{fig:sup_fig_3a}). The QA network is an LSTM initialized with the agent's internal state that we unroll for a fixed number of steps feeding the question as input at each step. We found that, indeed, the answering accuracy increased with the number of unroll steps from 1 to 12, while greater number of steps became detrimental. We performed the same analysis on the LSTM agent and found that regardless of the capacity of the QA network, we could not decode the correct answer from its internal state, suggesting that the limiting factor is not the capacity of the decoder but the lack of useful representations in the LSTM agent state.

\subsection{Answering accuracy during training for all questions}

\begin{figure*}[h]
    \centering
    \includegraphics[width=0.9\textwidth]{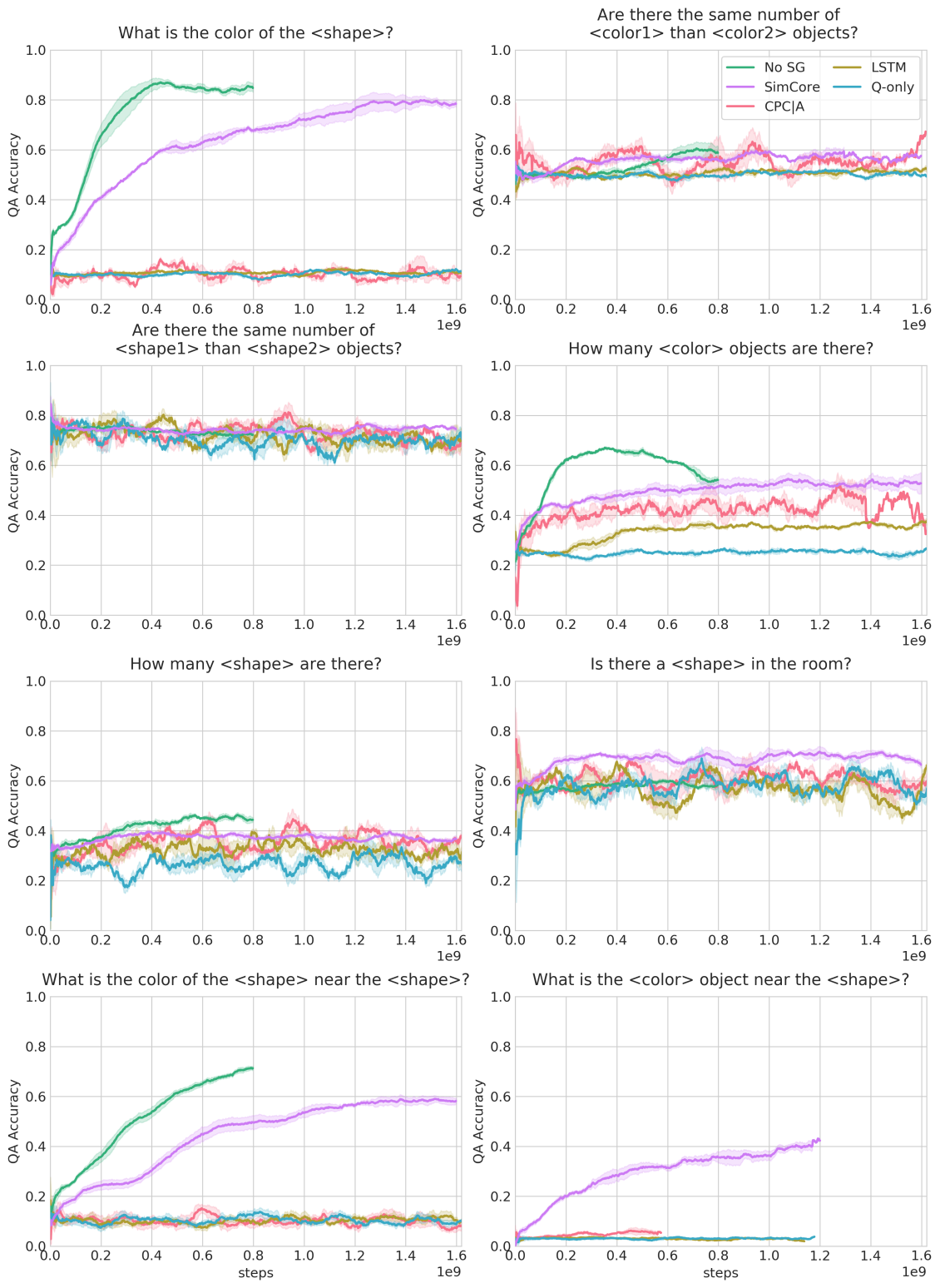}
    \caption{QA accuracy over training for all questions and all models.}
    \label{fig:sup_fig_3b}
\end{figure*}

The QA accuracy over training for all questions is shown in Figure~\ref{fig:sup_fig_3b}.

\subsection{Environment}
\label{sec:supp_environment}

Our environment is a single L-shaped $3$D room, procedurally populated with an assortment of objects.

\textbf{Actions and Observations.} The environment is episodic, and runs at $30$ frames per second. Each episode takes $30$ seconds (or $900$ steps). At each step, the environment provides the agent with two observations: a $96$x$72$ RGB image with the first-person view of the agent and the text containing the question.

The agent can interact with the environment by providing multiple simultaneous actions to control movement (forward/back, left/right), looking (up/down, left/right), picking up and manipulating objects (4 degrees of freedom: yaw, pitch, roll + movement along the axis between agent and object).

\textbf{Rewards.} To allow training using cross-entropy, as described in Section \ref{approach:qa}, the environment provides the ground-truth answer instead of the reward to the agent.

\textbf{Object creation and placement.}
We generate between $2$ and $20$ objects, depending on the task, with the type of the object, its color and size being uniformly sampled from the set described in Table \ref{table:appendix_env}.

Objects will be placed in a random location and random orientation. For some tasks, we required some additional constraints - for example, if the question is "What is the color of the cushion near the bed?", we need to ensure only one cushion is close to the bed. This was done by checking the constraints and regenerating the placement in case they were not satisfied.

\begin{table*}[h]
        \begin{tabular}{@{}lll@{}}
        \toprule
        Attribute & Options \\
        \midrule
        Object & basketball, cushion, carriage, train, grinder, candle, teddy, chair, \\
        & scissors, stool, book, football, rubber duck, glass, toothpaste, arm chair, \\
        & robot, hairdryer, cube block, bathtub, TV, plane, cuboid block, \\
        & car, tv cabinet, plate, soap, rocket, dining table, pillar block, \\
        & potted plant, boat, tennisball, tape dispenser, pencil, wash basin, \\
        & vase, picture frame, bottle, bed, helicopter, napkin, table lamp, \\
        & wardrobe, racket, keyboard, chest, bus, roof block, toilet \\ [0.05in]
        Color & aquamarine, blue, green, magenta, orange, purple, pink, red, \\
        & white, yellow \\ [0.05in]
        Size & small, medium, large \\ [0.05in]
        \bottomrule
        \end{tabular}
    \vspace{5pt}
    \caption{Randomization of objects in the Unity room.
        $50$ different types, $10$ different colors and $3$ different scales.}
    \label{table:appendix_env}
\end{table*}

\begin{table*}[h]
        \begin{tabular}{@{}lll@{}}
        \toprule
        \textbf{Body movement actions} & \textbf{Movement and grip actions} & \textbf{Object manipulation}  \\ \midrule
        NOOP                           & GRAB                               & GRAB + SPIN\_OBJECT\_RIGHT    \\
        MOVE\_FORWARD                  & GRAB + MOVE\_FORWARD               & GRAB + SPIN\_OBJECT\_LEFT     \\
        MOVE\_BACKWARD                 & GRAB + MOVE\_BACKWARD              & GRAB + SPIN\_OBJECT\_UP       \\
        MOVE\_RIGHT                    & GRAB + MOVE\_RIGHT                 & GRAB + SPIN\_OBJECT\_DOWN     \\
        MOVE\_LEFT                     & GRAB + MOVE\_BACKWARD              & GRAB + SPIN\_OBJECT\_FORWARD  \\
        LOOK\_RIGHT                    & GRAB + LOOK\_RIGHT                 & GRAB + SPIN\_OBJECT\_BACKWARD \\
        LOOK\_LEFT                     & GRAB + LOOK\_LEFT                  & GRAB + PUSH\_OBJECT\_AWAY     \\
        LOOK\_UP                       & GRAB + LOOK\_UP                    & GRAB + PULL\_OBJECT\_CLOSE    \\
        LOOK\_DOWN                     & GRAB + LOOK\_DOWN                  &                               \\ \bottomrule
        \end{tabular}
    \vspace{5pt}
    \caption{Environment action set.}
    \label{table:appendix_action_set}
\end{table*}

\end{document}